\documentclass[runningheads]{llncs}

\usepackage{eccv}

\usepackage{eccvabbrv}

\usepackage{graphicx}
\usepackage{booktabs}

\usepackage[accsupp]{axessibility}  

\usepackage{amsmath,amsfonts,bm}

\usepackage{algorithmic}
\usepackage{algorithm}

\def\eqref#1{equation~\ref{#1}}

\def\1{\bm{1}}

\def\vc{{\bm{c}}}

\def\vs{{\bm{s}}}

\def\vx{{\bm{x}}}

\def\mX{{\bm{X}}}

\DeclareMathAlphabet{\mathsfit}{\encodingdefault}{\sfdefault}{m}{sl}
\SetMathAlphabet{\mathsfit}{bold}{\encodingdefault}{\sfdefault}{bx}{n}

\usepackage{hyperref}
\usepackage{adjustbox}
\usepackage{multirow}
\usepackage{wrapfig}
\usepackage{booktabs}
\usepackage{pifont}

\usepackage{orcidlink}

\begin{document}

\title{ClusterStyle: Modeling Intra-Style Diversity with Prototypical Clustering for Stylized Motion Generation} 

\titlerunning{ClusterStyle}

\author{Kerui Chen\inst{1*}\orcidlink{0009-0007-6990-8833} \and
Jianrong Zhang\inst{1*} \and
Ming Li\inst{2} \and \\
Zhonglong Zheng\inst{3} \and
Hehe Fan\inst{1}$^{\dagger}$}

\authorrunning{K. Chen et al.}

\institute{CCAI, Zhejiang University, China \and
School of Artificial Intelligence, The Chinese University of Hong Kong, Shenzhen, China \and
Zhejiang Normal University, China}

\maketitle
\def\thefootnote{*}\footnotetext{Equal Contribution.}
\def\thefootnote{$\dagger$}\footnotetext{Corresponding author.}

\begin{abstract}
Existing stylized motion generation models have shown their remarkable ability to understand specific style information from the style motion, and insert it into the content motion. However, capturing intra-style diversity, where a single style should correspond to diverse motion variations, remains a significant challenge. In this paper, we propose a clustering-based framework, \textbf{ClusterStyle}, to address this limitation. 
Instead of learning an unstructured embedding from each style motion, we leverage a set of prototypes to effectively model diverse style patterns across motions belonging to the same style category. 
We consider two types of style diversity: global-level diversity among style motions of the same category, and local-level diversity within the temporal dynamics of motion sequences. These components jointly shape two structured style embedding spaces, \ie, global and local, optimized via alignment with non-learnable prototype anchors. Furthermore, we augment the pretrained text-to-motion generation model with the Stylistic Modulation Adapter (SMA) to integrate the style features. Extensive experiments demonstrate that our approach outperforms existing state-of-the-art models in stylized motion generation and motion style transfer. Project page: \url{https://1233chen.github.io/ClusterStyle/}.
\end{abstract}    
\section{Introduction}
\label{sec:intro}
\begin{figure}[t]
\centering
\includegraphics[width=0.8\linewidth]{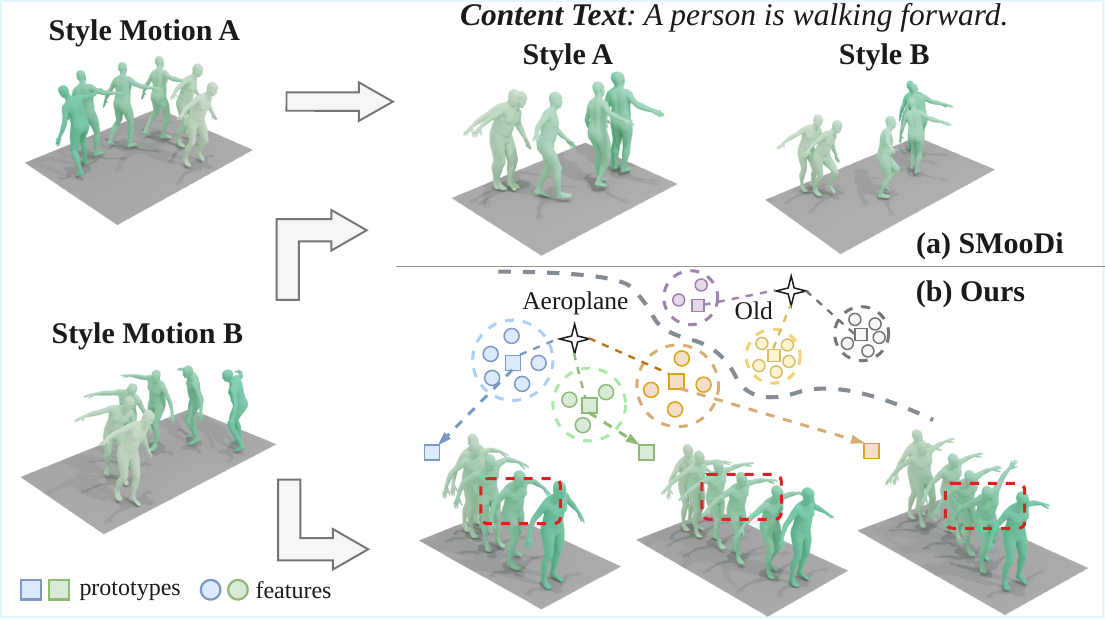}
\vspace{-2mm}
\caption{Comparison between SMooDi~\cite{zhong2024smoodi} and ClusterStyle. Given two style motions with the same style but differing in expression, along with a content text, (a) SMooDi generates similar stylized motions. Meanwhile, the generated motions still reflect elements from the style motions' content, leading to inconsistency with the intended content text.
(b) In contrast, our method uses different \textbf{prototypes} of the areoplane style, generating diverse motion with varying motion magnitudes (highlighted in \textcolor{red}{red}), while maintaining stronger alignment with the content semantics.}
\label{fig:intro}
\vspace{-3mm}
\end{figure}

Stylization is an essential technique for generative modeling~\cite{gatys2016image, ye2023ip}, enabling the rendering of source content in a target style. In the stylized motion generation~\cite{tao2022style, mason2022real, tang2023rsmt}, the primary aim is to transfer the style from a reference motion sequence to a source motion sequence while preserving the original content, which makes it useful in various real-world applications, including augmented reality~\cite{tao2022style}, animation~\cite{aberman2020unpaired}, and virtual avatars~\cite{ahuja2021coolmoves}.

Motion style is inherently expressive~\cite{motion_style_survey, zhong2025smoogpt}, where the same style can exhibit diverse motion variations influenced by the human's mood, intention, or context. For example, given a textual description ``a person is walking forward", the ``aeroplane" style should encompass a wide range of postures as well as variations in action amplitude, rather than a single rigid walk with both arms raised.

Recently, many works aim to achieve stylized motion generation and motion style transfer using diffusion models~\cite{hu2024diffusion, zhong2024smoodi, li2024mulsmo, guo2025stylemotif, qian2025mafd, raab2024monkey, sawdayee2025dance, song2024arbitrary}.
Song et al.~\cite{song2024arbitrary} leveraged trajectory as an auxiliary condition during the denoising process to improve the preservation of the content.
SMooDi~\cite{zhong2024smoodi} introduces a large-scale dataset for stylized motion generation from both the textual description and style motion. Building on the pretrained text-to-motion model, \ie, MLD~\cite{chen2023executing}, it uses a ControlNet-like architecture to serve as the style encoder.
BiFlow~\cite{li2024mulsmo} further replaces the style-to-content flow with a bidirectional control flow to reduce conflicts between content and style.

However, these methods embed the style motion into an \textit{unstructured} feature, which often cannot model the intra-style diversity, resulting in relatively uniform motion output. Meanwhile, we find that such unstructured features include both content and style information of the style motion, and the content part may lead to deviations from the target content intended by the text. For example, in \cref{fig:intro}(a), given two style motions with the same style but to different extents, and a text prompt ``a person is walking forward'', SMooDi generates relatively uniform motions, and does not reflect ``walking forward''.

In this paper, we propose ClusterStyle to address these limitations. It is inspired by recent clustering-based representation learning methods~\cite{zhou2022rethinking,liang2023clustseg} in computer vision, where cluster centers are progressively updated through interactions with pixel features. Specifically, each style category is represented by clustering its style motions into $K$ non-learnable prototypes (\ie, distinct centroids). Then, we build a transformer-based architecture as the style motion encoder, and the extracted features are used to update the prototypes iteratively during training. To facilitate the learning of a high-quality and disentangled style feature space, we encourage the contrastive property across prototypes belonging to different categories (Prototype-Based Inter-Style Learning) as well as among distinct prototypes within the same category (Prototype-Based Intra-Style Learning). Furthermore, we observe that a single motion sequence may also exhibit stylistic differences over time. Thus, we propose hierarchical clustering, which performs clustering on the entire motion sequences (\textit{global-level}) as well as their temporal segments (\textit{local-level}). Finally, we present a Stylistic Modulation Adapter (SMA) to effectively integrate the style information into the text-to-motion model.

Our approach characterizes two appealing advantages.
First, we explicitly model the intra-style diversity through learning multiple clustering centroids on both global and local levels of motion sequences. 
As shown in \cref{fig:intro}(b), using different cluster centroids (prototypes), our method can generate expressions of the same style to varying extents, making ClusterStyle a flexible and \textit{interpretable} framework.
Second, prototypes focus more on style attributes, enabling a better consistency between the motion and the target content described by the text. We evaluate the proposed approach on the 100STYLE~\cite{zhong2024smoodi} and HumanML3D~\cite{guo2022generating} datasets across two tasks, \ie stylized motion generation and motion style transfer. Extensive experiments demonstrate that ClusterStyle outperforms current state-of-the-art methods in fidelity and content preservation.

Our contributions can be summarized as follows: \textbf{1.} We propose ClusterStyle, a clustering-based framework that facilitates diverse stylized motion generation. \textbf{2.} We propose to use multiple clustering-based prototypes to model the intra-style diversity and consider both global-level and local-level discrepancies. \textbf{3.} We introduce Stylistic Modulation Adapter (SMA) for style attribute injection.
\section{Related Work}
\label{sec:related}

\noindent\textbf{Text-to-Motion Generation.}
The task of generating human motion from textual descriptions has seen rapid advancement in recent years. Early studies primarily focus on constructing a shared embedding space for text and motion~\cite{ahuja2019language2pose,ghosh2021synthesis,tevet2022motionclip,petrovich2022temos,petrovich21actor,petrovich23tmr, cheng2026unisonharmonizingmotionspeech}. 
Further advancements introduce the idea of discretizing the motion representation through vector quantization~\cite{guo2022tm2t}. 
For instance, T2M-GPT~\cite{zhang2023generating} presents VQ-VAE and GPT along with some training recipes, \eg, corruption strategy, which significantly improve the performance on the large-scale dataset, \ie, HumanML3D~\cite{guo2022generating}. Based on this line, several works improve this paradigm by enhancing the VQ-VAE~\cite{guo2024momask,Zhong_2023_ICCV} and leveraging masked token modeling~\cite{mmm,bamm, guo2024momask}.

Recently, diffusion models have become an increasingly popular choice for motion generation. The foundation of the diffusion model in this domain is laid by MDM~\cite{tevet2022human} and MotionDiffuse~\cite{zhang2024motiondiffuse}. They both use the Transformer-based denoising network to progressively refine random noises to synthesize human motions guided by textual descriptions.
Subsequent works~\cite{dabral2023mofusion, wei2025acmo, xie2024towards, wang2023fg, jin2023act, goel2024iterative} introduce diverse additional information to improve the quality or controllability of motion generation, such as spatial constraints~\cite{xie2023omnicontrol, wan2024tlcontrol, karunratanakul2023guided}, physical constraints~\cite{yuan2023physdiff} and multimodal information~\cite{zhang2023remodiffuse,zhou2023ude, han2024amd, chen2026scaling, chen2025prompt}. MLD~\cite{chen2023executing} proposes to shift the diffusion process to the latent space. The following efforts have been dedicated to improving performance~\cite{gao2024guess,zhang2025energymogen} and sampling efficiency~\cite{zhang2024motion} through architectural refinement.
Motion latent diffusion model is more relevant to our approach, as the current stylized motion generation methods are built upon MLD~\cite{chen2023executing}. We augment this architecture with our designed Stylistic Modulation Adapter (SMA) to effectively fuse text and style.

\noindent\textbf{Stylized Motion Generation.}
Stylized motion generation aims to synthesize human motions that reflect a desired content and style, which can be specified through text~\cite{sawdayee2025dance, kim2025personabooth} or reference motion~\cite{tao2022style, mason2022real, park2021diverse, tang2023rsmt, wen2021autoregressive, xu2020hierarchical, qian2025mafd, li2026vla, He2023UnifiedSGGHOI, he2022towards}.
Earlier approaches \cite{aberman2020unpaired,jang2022motion} primarily focused on motion style transfer and employed autoencoder-based architectures with AdaIN \cite{huang2017arbitrary} to disentangle and recombine content and style features.
Following this, Guo \cite{guo2024generative} explored style transfer in a probabilistic latent space, while MOST \cite{kim2024most} improves the recombination technique between the content and style.
Recently, diffusion-based methods~\cite{song2024arbitrary, raab2024monkey, hu2024diffusion} have increasingly been adopted in this task, primarily leveraging the prior knowledge from pre-trained text-to-motion models.
For instance, Hu et al.~\cite{hu2024diffusion} treat the denoising process as a style transfer stage and utilize CLIP~\cite{radford2021learning} for transfer guidance, and MoMo~\cite{raab2024monkey} achieves zero-shot style transfer by exploring the attention mechanism.

To facilitate research in this area, SMooDi \cite{zhong2024smoodi} proposes a new dataset named 100STYLE, which is paired with content text and style motion for a style-specific task.
In addition to introducing a new dataset, SMooDi proposed a ControlNet-like architecture to incorporate style features into the generation process.
To mitigate conflicts between content and style, BiFlow~\cite{li2024mulsmo} refines the style-to-content flow in the SMooDi into a bidirectional control flow.
StyleMotif~\cite{guo2025stylemotif} leverages multi-modal style inputs (e.g., text, audio, motion) and proposes a style-content cross-fusion mechanism to effectively integrate this information.

However, these methods all neglect the importance of intra-style diversity, leading to the lack of diversity and distinctiveness in the generated results when guided by different style motions.
In this paper, we represent each style category with multiple prototypes, obtained by clustering style embeddings from the full style motion dataset, which are used to model the intra-diversity of style.

\noindent\textbf{Clustering in Vision.}
Clustering enables models to automatically mine underlying patterns and learn meaningful representations from data.
Early clustering methods \cite{reynolds2015gaussian, jolion1991robust, frigui2002robust} primarily relied on raw data representations and hand-crafted priors, which were not effective when faced with complex data. Some methods~\cite{caron2020unsupervised, caron2018deep, yan2020clusterfit,ding2023decoupling} propose to combine deep learning techniques with prototype-based clustering. For example, Wang et al.~\cite{wang2022visual} considered the representation learning from the clustering perspective, constructing multiple cluster centers for each class to improve the recognition ability of the model.
This clustering paradigm has inspired a broad spectrum of research across domains,
such as image segmentation~\cite{zhou2022rethinking, liang2023clustseg, ding2024clustering, liang2023clusterfomer, niu2022spice, chen2024neural,zhang2022region}, point cloud analysis~\cite{feng2023clustering, liu2023point, feng2024interpretable3d, liu2024pointcluster}, and zero-shot learning~\cite{lu2024zero, qu2025learning}.

Inspired by these methods, we extend the clustering technique to stylized motion generation. Different from previous works that focus on global information of each data instance (\ie, image), ClusterStyle considers both long-term and short-term relationships within style motion sequences, facilitating a more fine-grained understanding of underlying style attributes and expression patterns.

\section{Method}
    \begin{figure}[t]
    \centering
    \includegraphics[width=0.95\linewidth]{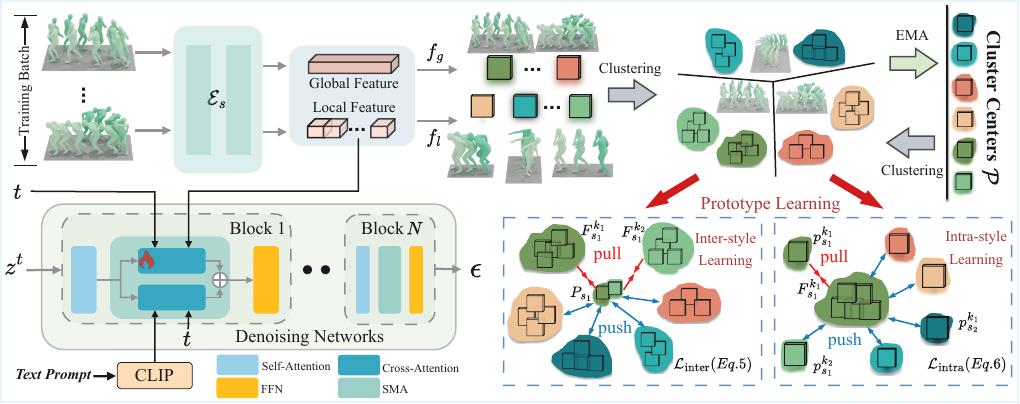}
    \caption{The overview of the ClusterStyle. Our method consists of a style encoder and a motion latent diffusion model. In the style encoder, we present a cluster-based prototype learning paradigm that represents each style category using a set of non-learnable prototypes (cluster centers) to model the intra-style diversity explicitly. Then, two contrastive losses are proposed for prototype-based intra-style learning and inter-style learning, respectively. To incorporate the style embedding into the diffusion process, we introduce a Style Modulation Adapter (SMA), enabling effective guidance of stylized motion generation.}
    \label{fig:method}
\end{figure}

In this paper, we aim to model intra-style diversity for stylized 3D motion generation.
As illustrated in \cref{fig:method}, we introduce ClusterStyle, a novel framework that comprises a motion latent diffusion model and a style encoder with a cluster-based prototype learning strategy.
\cref{sec3.1} outlines the problem formulation and provides an overview of our approach.
We introduce cluster-based style prototype learning, hierarchical style modeling, and style modulation adapter in \cref{sec3.2},~\cref{sec3.3}, and~\cref{sec3.4}, respectively.
Finally, implementation details are provided in \cref{sec3.5}.

\subsection{Problem Setting and Overview}
\label{sec3.1}

Given a style motion sequence $\mX_s = [\vx_\vs^1; \vx_s^2; \cdots; \vx_s^{L_s}]$ and a content text $c$, where $\vx_s^i \in \mathbb{R}^{d}$, $L_s$ and $d$ are the length and dimension of style motion. Our goal is to generate a motion sequence $\mX_c = [\vx_c^1; \vx_c^2; \cdots; \vx_c^{L_c}]$ with length $L_c$ that is consistent with the text while adhering to the style demonstrated by the style motion $\mX_s$.

Specifically, following previous works~\cite{zhong2024smoodi,li2024mulsmo, guo2025stylemotif}, we build our approach on a motion latent diffusion model~\cite{chen2023executing}, which consists of a VAE and latent diffusion model, pretrained on the HumanML3D dataset~\cite{guo2022generating}. During training, with the style motion-text pair $\{\mX_s,c_s\}$ available, we first feed the style motion into the VAE encoder to obtain the latent feature $z_s$. We gradually add Gaussian noise $\epsilon$ to $z_s$ over $T$ times via $z^t_s = \sqrt{\alpha^t} z_s^0 + \sqrt{1-\alpha^t} \epsilon$, where $t$ is the timestep and $\{\alpha^t\}_{t=0}^T$ is the noise variance schedule. Then, we propose a clustering-based style encoder which uses a set of prototypes to represent a single style class for intra-style diversity (\cref{sec3.2}). It also captures style information hierarchically by modeling both global-level motion and local-level temporal segments (\cref{sec3.3}), obtaining the style embedding $f_s$.
Then, we train a denoising autoencoder to predict the noise conditioned on $f_s$ and $c_s$, the loss function can be formulated as:
\begin{equation}
\mathcal{L}_{\text{diff}}^s = \mathbb{E}_{z_s, \epsilon, t, c_s, f_s}\Big[ \Vert \epsilon - \epsilon_\theta(z^t_s, t, c_s, f_s)) \Vert_{2}^{2}\Big].
\end{equation}

Furthermore, to prevent forgetting content-related prior knowledge, we also train the diffusion model on the content motion-text pair $\{\mX_c, c\}$ from the HumanML3D dataset,
\begin{equation}
\mathcal{L}_{\text{diff}}^c = \mathbb{E}_{z_c, \epsilon, t, c, f_c}\Big[ \Vert \epsilon - \epsilon_\theta(z^t_c, t, c, f_c)) \Vert_{2}^{2}\Big],
\end{equation}
where $z_c$ and $f_c$ denote the motion latent features from the VAE encoder and the style encoder, respectively. Different from existing methods~\cite{zhong2024smoodi} that employ a ControlNet-like~\cite{zhang2023adding} architecture for style feature fusion, we utilize a cross-attention to understand the textual description~\cite{zhang2025energymogen, zhang2026towards} and further propose a Style Modulation Adapter (\cref{sec3.4}) for style feature integration.

Combining these components, we optimize ClusterStyle using the following overall training objective:
\begin{equation}
    \mathcal{L_{\text{total}}} = \mathcal{L}_{\text{diff}}^s + \mathcal{L}_{\text{diff}}^c + \lambda_{\text{style}}\mathcal{L}_{\text{style}},
\end{equation}
where $\lambda$ is a hyper-parameter to balance the weight of $\mathcal{L}_{\text{style}}$, which supervises the training of the style encoder (see \cref{sec3.3} for more details).

\subsection{Clustering-based Style Prototype Learning}
\label{sec3.2}
The design of ClusterStyle is motivated by a key insight: motion style is inherently diverse, a single style should not be confined to a fixed form, but can manifest through multiple dynamic patterns. To achieve this, we raise two questions: \ding{182} \textit{How to model the intra-style diversity?} \ding{183} \textit{How can intra-style diversity be automatically mined without relying on manual annotation?} To answer these questions, we propose a clustering-based framework for style prototype learning, which enables the discovery of underlying sub-style patterns within each style category.

\noindent\textbf{Cluster Center Initalization.}
As a response to question \ding{182}, we propose to represent each style category using multiple prototypes. Each prototype captures a distinct sub-style pattern and acts as a reference point that encourages variation within the same style. This helps organize the style features into a well-structured and diverse style space. Specifically, for a style category $\vs$, we assume it contains $K_g$ sub-style patterns and define a cluster using a set of prototypes (\textit{a.k.a.} cluster centers) $P_s = \{p_s^i\}_{i=1}^{K_g}$, where $p_s^i \in \mathbb{R}^{1 \times d'}$ and $d'$ is the dimension of the prototype. We encode the style motion feature using a transformer-based style encoder $\mathcal{E}_s$, which can be computed as $f_g = \mathcal{E}_s(\mX_s)$. 
Our goal is to map the style feature $f_g$ to the closest prototype within $P_s$, which represents the corresponding style cluster.
To extend this idea, the prototype clusters for all style categories are collectively defined as $\mathcal{P} = \{P_s\}_{s=1}^{S}$, where $S$ is the total number of style categories.

\noindent\textbf{Prototype Assignment.}
It is challenging to directly supervise the assignment of features to prototypes due to the lack of manual annotations that specify or differentiate these prototypes. Driven by question \ding{183}, we formulate this process as an unsupervised optimal transport problem.
Formally, given a series of style motion features $F_s = [f_{g,s}^1, f_{g,s}^2, \cdots, f_{g,s}^{N_s}] \in \mathbb{R}^{d' \times N_s}$, where $N_s$ denotes the total motion numbers of the style category $s$. Correspondingly, we define a prototype matrix $P'_s \in \mathbb{R}^{d' \times K_g}$ representing $K_g$ style-specific prototypes. Note that each column of $F_s$ and $P'_s$ is L2-normalized. Based on a binary assignment matrix $L_s\in {\{0,1\}}^{K_g \times N_s}$, we compute the prototype assignment map as $A_s ={P'_s }L_s \in \mathbb{R}^{d' \times N_s}$, where each column of $A_s$ represents the aggregated prototype assigned to the corresponding style motion feature. We measure the inner product similarity $\langle \cdot, \cdot \rangle_I$ and maximize $\langle A_s, F_s\rangle_I$ to determine $L_s$. To prevent all style motion features from being assigned to a single prototype, we introduce a balancing constraint that encourages each prototype to be selected approximately $\frac{N_s}{K_g}$ times on average. We adopt the Sinkhorn algorithm~\cite{cuturi2013sinkhorn} to solve this problem, which introduces entropy regularization to enable fast and stable computation of the transport plan:
\begin{equation}
\max_{L_s}\langle A_s,F_s\rangle_I+\mu~\text{ KL }(L_s||\frac{\bm{1}}{{K_g} {N_s}}\bm{1}_{K_g}\bm{1}_{N_s}^{\top}),
\end{equation}
where KL is used to smooth the distribution via a hyper-parameter $\mu$. There are two additional constraints imposed on relaxed $L_s$: (1) $\bm{L}_s\in \mathbb{R}_+^{K_g\times N_s}$ ensuring that each feature is assigned once; and (2) $ {\bm{L}}_{s}\bm{1}_{N_s}=\frac{N_s}{K_g}\bm{1}_{K_g}$, encouraging balanced usage of prototypes.

\noindent\textbf{Prototype-based Contrastive Learning.} 
After tackling questions \ding{182} and \ding{183}, the next step aims to learn the high-quality representation of $\mathcal{P}$ and style motions. We propose two prototype-based contrastive losses $\mathcal{L}^g_{\text{inter}}$ and $\mathcal{L}^g_{\text{intra}}$, corresponding to Prototype-based Inter-Style Learning and Prototype-based Intra-style Learning, respectively. Specifically, with the style motion feature $f_g$, the prototype-based inter-style learning is designed to enhance inter-style discrimination by
pulling each motion feature closer to the prototype of its style category and pushing it away from prototypes of all other styles. The inter-style contrastive loss $\mathcal{L}^g_{\text{inter}}$ is formulated as follows:
\begin{equation}
    \mathcal{L}^g_{\text{inter}} = -\text{log} \frac{\text{exp}(-\text{sim}[{f_g,s}])}{\Sigma_{s'=1}^S \text{exp}(-\text{sim}[{f_g,s'}])},
\label{eq5}
\end{equation}
where $\text{sim}[{f_g,s}] =\min\{\cos(f_g, p_s^k) \}^{K_g}_{k=1}$ is a function to measure the similarity between the style feature $f_g$ and $P_s$, $\cos(\cdot)$ denotes the cosine similarity. We provide an ablation study in the Supplement to evaluate the impact of different similarity metrics on model performance.

In prototype-based intra-style learning, each motion feature is contrasted against all prototypes by treating the assigned prototype as positive, and all others, even within the same style category, as negatives. Given the style motion feature $f_g$, the corresponding prototype $p_s^k$, we define the negative prototypes as $\hat{P}$, which consists of all prototypes in $\mathcal{P}$ except the assigned $p_s^k$. Then, the intra-style contrastive loss $\mathcal{L}^g_{\text{intra}}$ can be written as:
\begin{equation}
\small
    \mathcal{L}^g_{\text{intra}} = -\text{log}{\frac{\text{exp}(\cos({f_g},p_s^{k})/\tau)}{\text{exp}(\cos ({f_g},p_s^{k})/\tau) + \Sigma_{\hat{p}\in \hat{P}} \beta \ \text{exp}(\cos ({f_g},\hat{p})/\tau)}},
\label{eq_intra}
\end{equation}
where $\beta = 1 \text{ if } \hat{p} \not\sim f_g, \text{ else } 5$, $\not\sim$ indicates that $\hat{p}$ and $f_g$ are from different style categories, $\tau$ is a temperature hyper-parameter which is set to 0.05 following~\cite{chen2020simple}. Then, the final training objective is defined as follows:
\begin{equation}
    \mathcal{L}^g_{\text{style}} = \mathcal{L}^g_{\text{inter}} + \mathcal{L}^g_{\text{intra}}.
\end{equation}

By doing so, the proposed two loss functions encourage the model to focus more on style-relevant patterns and to learn style representations that are disentangled from content information, thereby preventing the target content from being influenced by content information embedded in the style motion.

\noindent\textbf{Prototype Update.} Unlike traditional classifiers optimized by gradient descent, our prototypes are non-parametric and non-learnable statistics computed as the centroids of their assigned style-feature sets.
Concretely, for category $s$ and prototype $k$, let $\hat{f}^k_{g,s}$ denote the mean of features assigned to that prototype.
At each training iteration, we update the prototype $p_s^k$ by EMA via below formular:
\begin{equation}
    p_{s}^k \leftarrow \lambda_p p_{s}^k + (1-\lambda_p){\hat{f}^k_{g,s}},
\end{equation}
where $\lambda_p\in[0,1]$ is a momentum coefficient.

\noindent\textbf{Prototype-based Guidance.} The prototypes can serve as guiding signals to generate diverse stylized motion. Specifically, we define the prototype-based guidance function 
\begin{equation}
    \epsilon_\theta(z^t,t,c,s) = \epsilon_\theta(z^t,t,c,s) + \gamma_g\nabla_{z^t}G_g(z^t,t,p^k_s),
\label{eq9}
\end{equation}
where $G_p(z^t,t,p^k_s)=1-\text{cos}(f_g^{x^0}, p^k_s)$, $\gamma_g$ is the guidance weight.
During the inference, $z^t$ is iteratively optimized to approach the sub-style pattern associated with the target prototype $p^k_s$. Here, $z^t$ is the noisy latent at the timestep $t$, and we transform it to $z^0$ using the predict noise, which is then decoded into $x^0$ through $\mathcal{D}(z^0)$.
$f_g^{x^0}$ is the style feature of $x^0$, and $\text{cos}(\cdot)$ denotes the cosine similarity.
Please find more details in Supplement.

\subsection{Hierarchical Style Modeling}
\label{sec3.3}
It is worth noting that motion inherently exhibits temporal dynamics where a single motion sequence may involve stylistic variations over time. To capture a fine-grained understanding of style patterns within a motion sequence, we propose a hierarchical style modeling framework that independently clusters features at both the global level (entire motion sequences) and the local level (temporal segments). Specifically, as we mentioned in \cref{sec3.2}, we encode the style motion sequence $X_s$ into $f_m\in \mathbb{R}^{L_s \times d'}$, the global style embedding $f_g$ is processed by an average pooling operation over the temporal dimension. For the local style embedding, we first divide $f_m$ into $L_w$ non-overlapping temporal segments with a window size of $w$, where $L_s = L_w \times w$. Then we apply the same pooling operation within each segment, obtaining $f_l \in \mathbb{R}^{L_w \times d'}$. We perform an identical clustering-based learning that clusters the local-level features into $K_l$ local prototypes. A prototype-based contrastive loss, $\mathcal{L}^l_{style}$, is then computed to guide the learning of fine-grained style representations. The final style loss $\mathcal{L}_{style}$ is formulated as follows:
\begin{equation}
    \mathcal{L}_{\text{style}} = \mathcal{L}^g_{\text{style}} + \mathcal{L}^l_{\text{style}}.
    \label{eq:prototype_guidance}
\end{equation}
Finally, the overall style representation $f_s \in \mathbb{R}^{(L_w + 1) \times d'}$, which encapsulates both hierarchical information and intra-style diversity, is constructed by concatenating the global-level feature $f_g$ and local-level features $f_l$, and is subsequently used to control the style in the generation process. Note that the guidance mechanism (\cref{sec3.2}, \cref{eq9}) used in the global prototype approach can also be applied to local prototypes.

\subsection{Style Modulation Adapter}
\label{sec3.4}
With the style representation $f_s$ available, we design a Style Modulation Adapter (SMA) to effectively integrate $f_s$ into the pre-trained text-to-motion model, while preserving its prior knowledge.
As shown in \cref{fig:method}, the proposed SMA takes both text embedding $c$ and style representation $f_s$ as conditions. It employs two cross-attention modules for the content branch and style branch, respectively, enabling separate control over semantic content and stylistic rendering. Specifically, in the content branch, the text embedding is first computed with $O_{\text{content}} = \text{Softmax}(\frac{QK^\top}{\sqrt{d}})V$ to obtain the content feature, where $Q = XW^q_c$, $K = \vc W^k_c$, $V = \vc W^v_c$ with $W^{q,k,v}_c \in \mathbb{R}^{d \times d}$. As for the style branch, we use another cross-attention to model the interaction between $f_s$ and the denoised latent motion $z$, which can be calculated as $O_{\text{style}} = \text{Softmax}(\frac{Q(K')^\top}{\sqrt{d}})V'$, where $K' = f_s W^k_s$, $V' = f_s W^v_s$ with $W^{k,v}_s \in \mathbb{R}^{d' \times d}$. We then combine $O_{\text{content}}$ and $O_{\text{style}}$ together, leading to the output $O_{\text{SMA}}$
\begin{equation}
    O_{\text{SMA}} = O_{\text{content}} + \lambda O_{\text{style}},
\end{equation}
where $\lambda$ is a trainable parameter that modulates the balance between content and style.

\subsection{Implementation Details}
\label{sec3.5}
ClusterStyle adopts a frozen CLIP ViT-L/14 model as the text encoder. For the style encoder, we use 6 transformer layers with a dimension of 512. We set the global prototype number $K_g$ to 3, local prototype number $K_l$ to 30, and momentum coefficient $\lambda_p$ to 0.95. Following~\cite{zhang2025energymogen}, we use a pretrained VAE with 5 latent vectors. The denoising autoencoder consists of $N=9$ layers of transformer blocks with a dimension of $d=256$. During training, ClusterStyle is optimized using the AdamW optimizer with a batch size of 128, and the learning rate is set to 2e-5 with a linear warm-up period of 500 iterations, followed by 3000 iterations of training with a constant learning rate. We first train the style encoder for 1800 iterations, after which prototype updates are frozen during the remaining training process. 
Training our model takes about 1 hour on a single NVIDIA A6000 GPU. Note that only $W^{k,v}_s$ in the Style Modulation Adapter (SMA) and parameters in the style encoder are updated. During inference, we generate stylized motions over 50 steps using the DDIM strategy. Following SMooDi~\cite{zhong2024smoodi}, we use classifier-free guidance for content texts and classifier guidance for style motions. Please find more information corresponding to the inference in the supplementary material.

\section{Experiment}

\begin{table*}[t!]
\centering
\setlength{\tabcolsep}{10pt}
\caption{Quantitative comparisons of ClusterStyle with existing state-of-the-art methods on the stylized motion generation task. The best and second best results are \textbf{bold} and \underline{underlined}.}
\label{stylization}
\begin{adjustbox}{width=1.0\linewidth}
\begin{tabular}{l|@{\hspace{1mm}}c@{\hspace{1mm}}|cccccc}
\toprule
\multirow{2}{*}{Methods} & \multirow{2}{*}{Venue}
& \multirow{2}{*}{FID\ $\downarrow$}& \multirow{2}{*}{FSR $\downarrow$} & \multirow{2}{*}{MM Dist\ $\downarrow$} & {R-Precision} & \multirow{2}{*}{Diversity $\rightarrow$} & \multirow{2}{*}{SRA\ $\uparrow$} \\
&&&&& (Top-3) \ $\uparrow$&& \\
\midrule
Motion Puzzle~\cite{jang2022motion}&{\fontsize{7pt}{8.5pt}\selectfont TOG 2022}&6.127&0.185&6.467&0.290&6.576&63.769\\
Aberman~\cite{aberman2020unpaired}&{\fontsize{7pt}{8.5pt}\selectfont TOG 2020}&3.309&0.347&5.983&0.406&8.816&54.367\\
ChatGPT+MLD& - &0.614&0.131&4.313&0.605&8.836&4.819\\ 
SMooDi~\cite{zhong2024smoodi}&{\fontsize{7pt}{8.5pt}\selectfont ECCV 2024}&1.609&0.124&4.477&0.571&9.235&72.418\\
BiFlow~\cite{li2024mulsmo}&{\fontsize{7pt}{8.5pt}\selectfont ARXIV 2025}&\underline{1.527}&0.118&\underline{4.292}&\underline{0.613}&9.303&77.042\\
StyleMotif~\cite{guo2025stylemotif}&{\fontsize{7pt}{8.5pt}\selectfont ARXIV 2025}&1.551&\textbf{0.097}&4.354&0.586&7.567&\underline{77.650}\\
\midrule
ClusterStyle (Ours)&-&\textbf{1.137}&\underline{0.113}&\textbf{3.610}&\textbf{0.708}&8.719&\textbf{78.101} \\
\bottomrule
\end{tabular}
\end{adjustbox}
\end{table*}

\subsection{Experimental Settings}
We evaluate our approach on two tasks: stylized motion generation (\cref {sec4.2}) and motion style transfer (\cref{sec4.3}). We compare our model with five representative state-of-the-art methods: Aberman~\cite{aberman2020unpaired}, Motion Puzzle~\cite{jang2022motion}, SMooDi~\cite{zhong2024smoodi}, BiFlow~\cite{li2024mulsmo}, and StyleMotif~\cite{guo2025stylemotif}. Meanwhile, we also offer detailed analysis and discussion in \cref{sec4.4}. Please note that more information about datasets is provided in the supplementary material.

\noindent\textbf{Datasets.}
We conduct experiments using a single model on two datasets, \ie, HumanML3D~\cite{guo2022generating} and 100STYLE~\cite{zhong2024smoodi}.
HumanML3D \cite{guo2022generating} is used to preserve content-related prior knowledge.
100STYLE \cite{zhong2024smoodi} is used for providing a wide range of motion styles to guide style-specific learning.
During evaluation, content is derived from HumanML3D, while style references are taken from 100STYLE. Please see Supplement for more information.

\noindent\textbf{Evaluation Metrics.}
Following SMooDi~\cite{zhong2024smoodi}, we adopt standard evaluation metrics to comprehensively assess the quality of stylized motion. These metrics include: (1) R-Precision and Multi-modal Distance (MM-Dist) for content preservation; (2) Style Recognition Accuracy (SRA) for style fidelity; (3) Fréchet Inception Distance (FID) for motion quality; (4) \textit{Diversity} for motion diversity; (5) Foot Skating Ratio (FSR) for physical plausibility.

\begin{figure*}[t]
\centering
\includegraphics[width=0.93\linewidth]{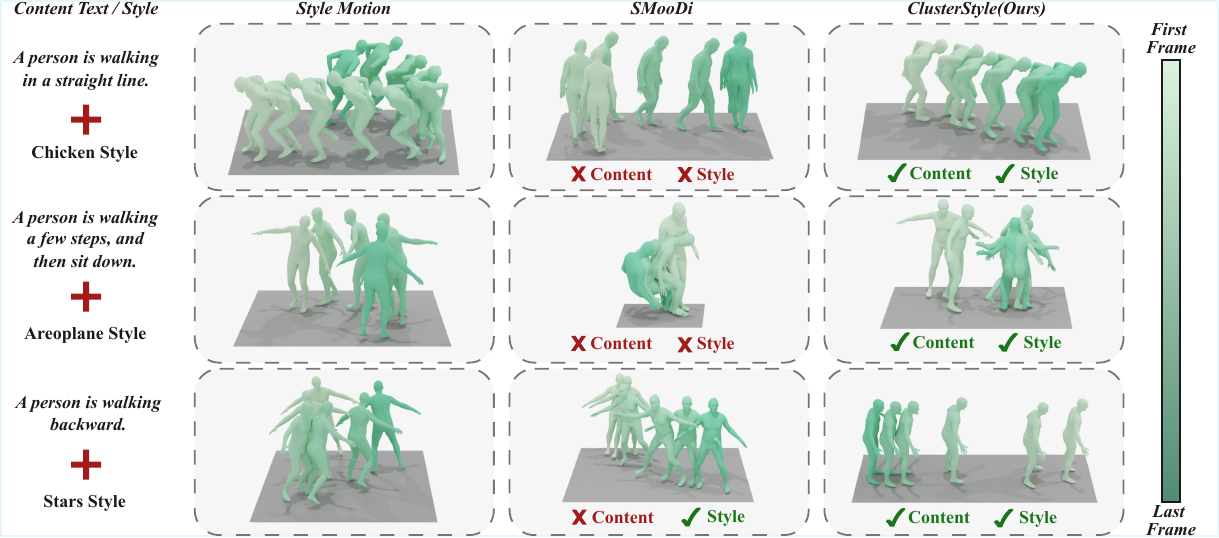}
\caption{Qualitative results of stylized motion generation. We compare our method with SMooDi under various text prompts and style motion inputs. Our approach demonstrates better content alignment and achieves more accurate and expressive style rendering. For example, the results of SMooDi are inconsistent with the motion trajectories (\eg, "backward", "straight") and action (e.g., "walk") described in the content. More visual comparisons can be found in the Supplement.}
\label{fig:vis1}
\end{figure*}

\subsection{Stylized Motion Generation}
\label{sec4.2}
\noindent\textbf{Quantitative Results.}
\cref{stylization} shows quantitative results on the HumanML3D and 100STYLE datasets. We compare our method with current state-of-the-art methods, \eg, SMooDi~\cite{zhong2024smoodi}, BiFlow~\cite{li2024mulsmo}, and StyleMotif~\cite{guo2025stylemotif}. Our method achieves R-Precision of 0.708 and SRA of 78.101, surpassing the BiFlow by 15.5\% and 1.1\%. In terms of FID, our method achieves a score of 1.137, representing a 27\% improvement over StyleMotif and demonstrating significantly improved generation fidelity. These results show that the motions generated by our approach are semantically aligned with the content text and stylistically consistent with the reference style motion.

\noindent\textbf{Qualitative Results.}
In \cref{fig:vis1}, we present visual comparisons between our model and baseline approach SMooDi~\cite{zhong2024smoodi}.
It can be seen that SMooDi struggles to retain essential motion details (\eg, “circular” or “walking” trajectories) and lacks fidelity in expressing target styles such as “aeroplane” or “chicken”.
Our approach is able to generate motions that better reflect the semantics of the text prompt, while also transferring correct stylistic characteristics from the reference motion.
These observations demonstrate the superior performance of our model, as well as its effectiveness in disentangling motion content and style.
Please find more visual results in the supplementary material.

\begin{table}[t]
  \centering
  \begin{minipage}[t]{0.42\textwidth}
    \centering
    \caption{Quantitative comparison with the state-of-the-art methods.}
    \begin{adjustbox}{width=\linewidth}
    \tabcolsep 2pt
    \footnotesize   
    \begin{tabular}{lcccc}
    \toprule
    {Method}&{FSR $\downarrow$}&{FID\ $\downarrow$}&{SRA\ $\uparrow$}\\
    \midrule
    Motion Puzzle~\cite{jang2022motion} & 0.197 & 6.871 & 67.233 \\
    Aberman~\cite{aberman2020unpaired} & 0.338 & 3.892 & 61.006 \\
    SMooDi~\cite{zhong2024smoodi} & 0.095 & 1.582 & 65.147 \\ 
    BiFlow~\cite{li2024mulsmo} & \underline{0.087} & 1.566 & \underline{70.238} \\
    StyleMotif~\cite{guo2025stylemotif} & 0.094 & \underline{1.375} & 68.810 \\
    ClusterStyle (Ours) & \textbf{0.078} & \textbf{0.768} & \textbf{73.849} \\
    \bottomrule
    \end{tabular}
    \end{adjustbox}
    \label{transfer}
  \end{minipage}
  \hfill
  \begin{minipage}[t]{0.54\textwidth}
    \caption{Ablation studies of different loss functions for the style encoder.}
    \begin{adjustbox}{width=\linewidth}
    \begin{tabular}{l|cccc}
    \toprule
    \footnotesize
    \multirow{2}{*}{\large $\mathcal{L}_{\text{style}}$}
    & \multirow{2}{*}{FID\ $\downarrow$}&R-Precision\ $\uparrow$ &{Diversity} & \multirow{2}{*}{SRA\ $\uparrow$} \\
    &&(Top-3) &$\rightarrow$&\\
    \cmidrule{1-5}
    $\mathcal{L}_{\text{inter}}$ & 1.112 & 0.712 & 8.479 & 75.182 \\ 
    $\mathcal{L}_{\text{intra}}$ & 1.349 & 0.673 & 8.509 & 67.746 \\ 
    $\mathcal{L}_{\text{entropy}}$ & 1.331 & 0.677 & 8.886 & 65.465 \\
    $\mathcal{L}_{\text{inter}} + \mathcal{L}_{\text{intra}}$ & 1.137 & 0.708 & 8.719 & 78.101 \\ 
    \bottomrule
    \end{tabular}
    \end{adjustbox}
    \label{loss_item}
  \end{minipage}
\end{table}

\begin{figure}[t]
  \centering
  \begin{minipage}[t]{0.55\textwidth}
        \centering
        \includegraphics[width=1.0\linewidth]{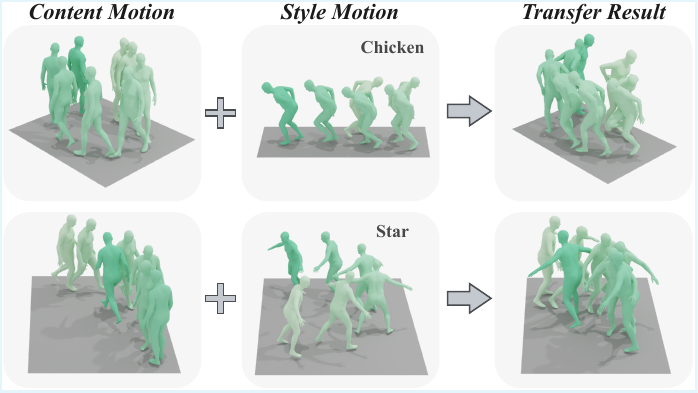}
        \caption{Qualitative results of motion style transfer. Our approach effectively transfers the target motion style, such as `Chicken’ or `Star’, onto the original motion, preserving its structure while adapting its stylistic characteristics.}
        \label{fig:transfer}
  \end{minipage}
  \hfill
  \begin{minipage}[t]{0.42\textwidth}
        \centering
        \includegraphics[width=0.97\linewidth]{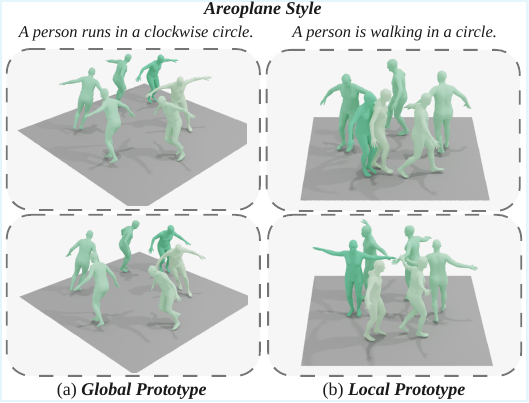}
        \caption{Visualization of prototype guiding. We visualize how global and local prototypes guide the stylization process for diverse generation results under the `Aeroplane' style.}
        \label{fig:diverse}
  \end{minipage}
\end{figure}

\subsection{Motion Style Transfer}
\label{sec4.3}
Our method utilizes SD-Edit \cite{meng2021sdedit} to achieve motion style transfer, relying solely on the pre-trained ClusterStyle model without additional fine-tuning. More implementation details and visualization results are available in the Supplement.

\noindent\textbf{Quantitative Results.}
\cref{transfer} presents the quantitative results, comparing our method against current state-of-the-art methods.
Our method achieves the lowest Foot Skating Ratio (0.078), reflecting a 17\% reduction compared to the best-performing method, \ie, StyleMotif, and indicating improved physical realism.
For FID and SRA, our methods achieve scores of 0.768 and 73.849, a 44\% and 5.03\% improvement over StyleMotif, demonstrating superior motion fidelity and transfer accuracy.

\noindent\textbf{Qualitative Results.}
\cref{fig:transfer} shows the visual results of ClusterStyle on motion style transfer. Given styles of ``chicken'' and ``star'', we observe that our method effectively preserves the action and trajectory of the content motion, while integrating styles consistent with style motions.

\begin{figure}[t]
    \centering
    \begin{minipage}[t]{0.55\textwidth}
        \vspace{-12pt}
        \centering
        \captionof{table}{Ablation studies of key components.}
        \vspace{8pt}
        \label{tab3}
        \begin{adjustbox}{width=\linewidth}
        \tabcolsep 2pt
        \footnotesize
        \renewcommand\arraystretch{0.95}
        \setlength{\tabcolsep}{4pt}
        \begin{tabular}{cc|ccccc}
            \hline
            \multirow{2}{*}{$K_g$} & \multirow{2}{*}{$K_l$} & \multirow{2}{*}{FID\ $\downarrow$}&\multirow{2}{*}{MM Dist\ $\downarrow$}&R-Precision\ $\uparrow$ &{Diversity} & \multirow{2}{*}{SRA\ $\uparrow$} \\
            &&&&(Top-3) &$\rightarrow$&\\
            \cmidrule{1-7}
            1 & 1 & 1.443 & 3.886 & 0.668 & 8.136 & 74.785 \\ \hline
            1 & 30 & 1.143 & 3.613 & 0.701 & 8.483 & 76.226  \\
            3 & 30 & \textbf{1.137} & \textbf{3.610} & \textbf{0.708} & 8.719 & 78.101 \\
            5 & 30 & 1.170 & 3.641 & 0.701 & 8.366 & 77.883 \\ 
            10 & 30 & 1.198 & 3.687 & 0.697 & 8.183 & \textbf{78.528} \\ \hline
            3 & 1 & 1.313 & 3.769 & 0.685 & 8.178 & \textbf{78.744} \\
            3 & 5 & 1.226 & 3.690 & 0.695 & 8.461 & 78.399 \\
            3 & 10 & 1.185 & 3.637 & 0.701 & 8.661 & 77.281 \\
            3 & 30 & \underline{1.137} & \underline{3.610} & \underline{0.708} & 8.719 & \underline{78.101} \\ 
            3 & 50 & \textbf{1.092} & \textbf{3.607} & \textbf{0.711} & 8.516 & 77.302 \\ \hline
        \end{tabular}
        \end{adjustbox}
    \end{minipage}
    \hfill
    \begin{minipage}[t]{0.42\textwidth}
        \vspace{0pt}
        \centering
        \includegraphics[width=\linewidth]{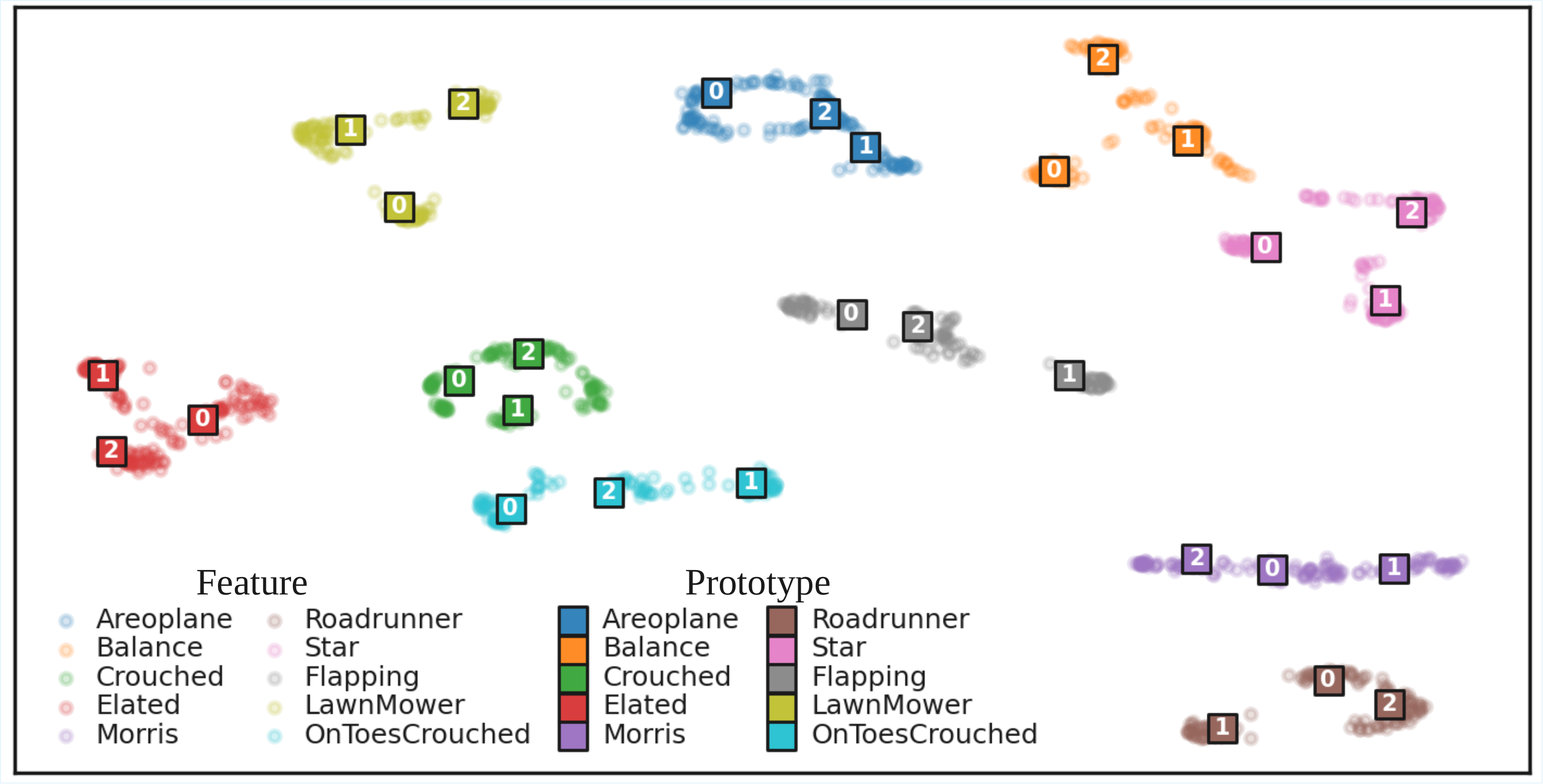}
        \captionof{figure}{(Please Zoom in for details.) Visualization of style feature space.}
        \label{fig:tsne}
    \end{minipage}
\end{figure}
\subsection{Discussion}
\label{sec4.4}
\noindent\textbf{Style Diversity.}
We explore the capability of our method to generate diverse motions from a single style.
In our experiments, we use different global prototypes as global style features to generate stylized motions, as shown in the left column of \cref{fig:diverse} and the bottom column of \cref{fig:intro}. Furthermore, we randomly combine local prototypes to generate stylized motions, as shown in the right column of \cref{fig:diverse}. We find that both different global prototypes and local prototypes yield stylistic results that vary in the extent of spread arms. Please note that, at inference time, our method enables diverse motion generation by integrating the style features obtained from the style encoder with different configurations of global and local prototypes. 

We also visualize the style feature space in \cref{fig:tsne} using t-SNE embeddings. We can observe that prototypes emerge as multiple distinct modes within each style. These diverse prototypes capture the intra-diversity and facilitate the generation of varied stylistic motions. Meanwhile, the visualization also reveals clear inter-style separation, ensuring that our model possesses the discriminative capability to handle similar style motions (\eg, \textit{Star} and \textit{Aeroplane}). 
To further validate this discriminative power, we provide an additional visual comparison in Fig.~\ref{fig:similar}. 
The primary distinction between \textit{Star} and \textit{Aeroplane} is the open-leg stance.
When conditioned on a \textit{Star} reference motion, SMooDi struggles to differentiate the two and produces Aeroplane-like results.
In contrast, our model accurately identifies the subtle discrepancies, consistently generating motions that remain faithful to the target style.
More results and analysis are provided in the supplementary material.

\noindent\textbf{Investigating Key Components.}
We first investigate the effectiveness of the clustering strategy and prototype number (\ie, $K_g$ and $K_l$) on the task of stylized motion generation, and the results are shown in \cref{tab3}.

\textit{Clustering Strategy:} We replace the clustering procedure with a learnable classification head, and the style loss $\mathcal{L}_{\text{style}}$ is replaced by a conventional cross-entropy loss. As indicated in the first row of \cref{tab3}, this modification leads to a 0.3 reduction in FID, a 5.6\% decrease in R-Precision, and a 3.32\% drop in SRA. This shows the effectiveness of the proposed cluster-based framework.

\textit{Global Prototype Number $K_g$:} The middle four rows of \cref{tab3} illustrate the impact of varying the number of global prototypes. $K_g=$3 yields relatively better results compared to other configurations. We observe a trade-off: as the number of global prototypes increases, content consistency metrics (e.g., R-Precision) tend to decline, whereas SRA improves. This suggests that while diversity benefits from a larger set of prototypes, content alignment may be negatively affected.

\textit{Local Prototype Number $K_l$:} The last five rows of \cref{tab3} show the impact of different number of local prototypes. $K_l=$30 achieves comparable performance to $K_l=$50, while being more computationally efficient. It can be seen that increasing local prototypes improves content preservation but degrades SRA, showing an opposite trend to global prototypes.

\noindent\textbf{Ablation Study of Training Objective.}
\cref{loss_item} presents the impact of different loss functions used in the style encoder.
Training with only inter-style $\mathcal{L}_{\text{inter}}$, $\mathcal{L}_{\text{intra}}$ or $\mathcal{L}_{\text{entropy}}$ leads to performance drops of 2.9\%, 10.4\%, and 12.6\% in SRA, respectively. Here, $\mathcal{L}{\text{entropy}}$ denotes a variant where learnable prototypes are trained using a standard cross-entropy loss instead of $\mathcal{L}_{\text{intra}}$. Our prototype-based contrastive learning achieves the best performance by combining $\mathcal{L}_{\text{inter}}$ and $\mathcal{L}_{\text{intra}}$ together.

\noindent\textbf{Impact of different similarity metrics.} We adopt cosine similarity in the prototype-based contrastive loss. As shown in \cref{L1_L2}, both alternatives lead to a decrease in SRA. In particular, adopting $L_1$ distance further degrades FID and Top-3 accuracy, indicating reduced motion quality and weaker alignment with the textual content.

\begin{table}[t]
    \centering
    \small
    \begin{minipage}[t]{0.63\textwidth}
        \vspace{0pt}
        \centering
        \includegraphics[width=\linewidth]{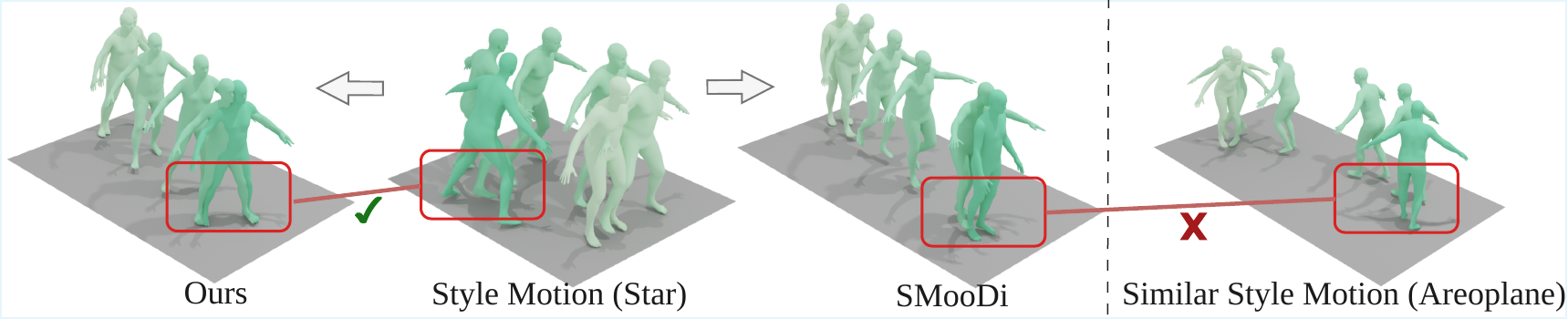}
        \vspace{0.1pt}
        \captionof{figure}{Qualitative comparison on easily confused Star style motion.} 
        \label{fig:similar}
    \end{minipage}
    \hfill
    \begin{minipage}[t]{0.33\textwidth}
        \vspace{0pt}
        \caption{Ablation study of different similarity metrics.}
        \label{L1_L2}
        \centering
        \begin{adjustbox}{width=1.0\linewidth}
        \tabcolsep 2pt
        \footnotesize   
        \begin{tabular}{l|cccccc}
        \toprule
        \multirow{2}{*}{\large $\text{sim}$}
        & \multirow{2}{*}{FID\ $\downarrow$}&\multirow{2}{*}{FSR\ $\downarrow$}&R-Precision\ $\uparrow$ &{Diversity} & \multirow{2}{*}{SRA\ $\uparrow$} \\
        &&&(Top-3) &$\rightarrow$&\\
        \midrule
        {$L_2$} & 1.191 & 0.114 & 0.706 & 8.553 & 76.483 \\
        {$L_1$} & 1.322 & 0.119 & 0.689 & 8.936 & 76.802 \\
        \text{cos} & 1.137 & 0.113 & 0.708 & 8.719 & 78.101\\
        \bottomrule
        \end{tabular}
        \end{adjustbox}
    \end{minipage}
\end{table}
\section{Conclusion}

In this paper, we propose ClusterStyle, a clustering-based framework for capturing intra-style diversity in stylized motion generation. Unlike prior methods that learn a single embedding per style, ClusterStyle represents style features using multiple clustering-based prototypes at both global and local levels. To fuse style features into the pretrained diffusion model, we introduce the Stylistic Modulation Adapter (SMA). Our approach achieves superior performance across stylized motion generation and motion style transfer. We conduct extensive ablation studies and provide visual results to validate the effectiveness of the proposed components. We also show that prototypes can learn diverse sub-style patterns with clear semantic meaning, enhancing both diversity and interpretability.

\clearpage
\section*{Acknowledgements}
This work was supported in part by the National Natural Science Foundation of China (92570101), the Major Program of the Natural Science Foundation of Zhejiang Province, China (LD26F020003) and the Earth System Big Data Platform of the School of Earth Sciences, Zhejiang University.

\bibliographystyle{splncs04}
\bibliography{main}
\end{document}